\documentclass[runningheads]{llncs}
\usepackage[utf8]{inputenc}
\usepackage{graphicx}
\usepackage{amsmath}
\usepackage{amsfonts}
\usepackage{mathtools}

\usepackage{booktabs}
\usepackage{tabularx}
\newcolumntype{C}{>{\centering\arraybackslash}X}
\newcolumntype{R}{>{\raggedleft\arraybackslash}X}

\newcommand\myeq{\mathrel{\overset{\makebox[0pt]{\mbox{\normalfont\tiny\sffamily def}}}{=}}}

\DeclareMathSizes{10}{9}{7}{5}
\let\subparagraph\relax
\usepackage[compact]{titlesec}
\titlespacing{\section}{0pt}{0.3cm}{0.2cm}
\titlespacing{\subsection}{0pt}{0.15cm}{0.1cm}
\titlespacing{\subsubsection}{0pt}{0.075cm}{0.1cm}

\begin{document}
\title{Efficient LSTM Training with Eligibility Traces}
%
%
\author{Michael Hoyer\inst{1}
\and
Shahram Eivazi\inst{2}
\and
Sebastian Otte\inst{1}%
}
\authorrunning{M. Hoyer et al.}
%
\institute{Neuro-Cognitive Modeling, University of Tübingen, Sand 14, 72076 Tübingen, Germany\\ 
\email{m-hoyer@gmx.de}, \email{sebastian.otte@uni-tuebingen.de}
\and
Autonomous Systems Lab, University of Tübingen, Sand 14, 72076 Tübingen, Germany\\
\email{shahram.eivazi@uni-tuebingen.de}
}
\maketitle              
\begin{abstract}
Training recurrent neural networks is predominantly achieved via backpropagation through time (BPTT). However, this algorithm is not an optimal solution from both a biological and computational perspective. A more efficient and biologically plausible alternative for BPTT is e-prop.  We investigate the applicability of e-prop to long short-term memorys (LSTMs), for both supervised and reinforcement learning (RL) tasks. We show that e-prop is a suitable optimization algorithm for LSTMs by comparing it to BPTT on two benchmarks for supervised learning. This proves that e-prop can achieve learning even for problems with long sequences of several hundred timesteps. We introduce extensions that improve the performance of e-prop, which can partially be applied to other network architectures. With the help of these extensions we show that, under certain conditions, e-prop can outperform BPTT for one of the two benchmarks for supervised learning. Finally, we deliver a proof of concept for the integration of e-prop to RL in the domain of deep recurrent Q-learning. 

\end{abstract}
\section{Introduction}

During the last decades, artificial neural networks (ANNs) have constantly pushed the boundaries of artificial intelligence further and further \cite{lecun2015deep}. While doing so, ANNs have diverged quite far in both architecture and ways of learning from their original source of inspiration: the human brain.

The long short-term memory (LSTM) \cite{hochreiter1997long}, for example, was not designed to model the recurrent networks of our brains, but specifically to learn long-term dependencies. The gold standard for training recurrent neural networks (RNNs) like LSTMs is backpropagation through time (BPTT) \cite{werbos1990backpropagation}. This method propagates errors backwards through the network and time to calculate error gradients with respect to specific weights. To do so, the whole network has to be unrolled before the error can be propagated backwards. This unrolling process is not only memory intensive, it makes effective online learning difficult and impractical. The longer a sequence is, the larger this problem becomes. BPTT views the optimization of networks purely from a mathematical perspective. There is no biological foundation BPTT is based on. In fact, we do not know how the connections between neurons in the brain are optimized. What we do know is that error backpropagation is most likely not the method used in the brain \cite{grossberg1987competitive}. The optimization of the highly recurrent networks of our brains must therefore be done differently. This raises hopes that such methods can be applied to train RNNs more efficiently. 

One biologically inspired method here is e-prop \cite{bellec2019biologically,bellec2020solution}. This method works by refactoring the equation of BPTT so that gradients can be computed as a combination of eligibility traces, computed forward in time, and online learning signals, estimating the loss \cite{bellec2019biologically,bellec2020solution}. This alleviates the needs for error backpropagation and unrolling of networks which makes this algorithm far more efficient than BPTT. E-prop was originally designed for spiking neural networks but can also be applied to LSTMs.


We apply e-prop to LSTMs and introduce possible extensions to increase it's performance and stability. To evaluate the algorithm and the extensions, we compare the performance for supervised learning tasks with BPTT. Furthermore we deliver a proof of concept that e-prop can be integrated into recurrent Q-learning and compare this approach to a BPTT based one. Considering real world problems are often a partially observable Markov decision process (POMDP). One possible solution for dealing with POMDPs is to incorporate RNNs into deep Q-learning \cite{hausknecht2015deep}. However, the same drawbacks as in supervised learning persist, which is a problem since RL tasks can take several hundred timesteps to be solved. 

\section{E-Prop}

E-prop calculates gradients in forward manner and hence is far more efficient than BPTT. This is achieved by refactoring the gradient computation from BPTT,
\begin{equation}
    \label{equation:generalequation}
    \frac{dE}{dw_{jk}} = \sum_t \frac{dE}{dz_k^t} \Biggl[\frac{dz_k^t}{dw_{jk}}  \Biggl]_{\textrm{local}} .
\end{equation}
The local gradient in this equation is not an approximation, it collects the maximal amount of information about $\frac{dE}{dw_{jk}}$ that can be computed in forward manner. For e-prop it is defined as the eligibility trace
\begin{equation} 
    e^t_{jk} \myeq \Biggl[\frac{dz_k^t}{dw_{jk}} \Biggl]_{\textrm{local}} .
\end{equation}
The other part of equation \ref{equation:generalequation}, $\frac{dE}{dz^t_k}$, is replaced by an approximation of the loss gradient, the online learning signal $l^t_k$. The computation of gradients with e-prop is therefore a combination of learning signals and eligibility traces
\begin{equation}
\label{equation:gradient_all_timesteps}
    \frac{dE}{dw_{jk}} = \sum_j e^t_{jk} l_k^t .
\end{equation}

\subsubsection{Eligibility Traces for LSTMs}

When applying e-prop to LSTMs, there are several aspects that need to be considered. Firstly, following \cite{bellec2020solution}, we use the cell state of the LSTM $c_k^t$ as hidden state for e-prop. Secondly, since LSTMs have individual weight matrices for the individual gates for both input and recurrent connections, individual eligibility traces need to be computed. We introduce $A \in \{i,f,\tilde{c}\}$ for input gate, forget gate and cell state candidate and $B \in \{\textrm{in}, \textrm{rec}\}$ for the connection type to specify an eligibility trace. A specific eligibility trace $e^{(A,B),t}_{jk}$ is then given by
\begin{equation}
    e^{(A,B),t}_{jk} = \epsilon_{jk}^{(A,B),t} \frac{\partial z_k^t}{\partial c_k^t} ,
\end{equation}
where $\epsilon_{jk}^{(A,B),t}$ denotes a specific eligibility value of timestep $t$. The derivative in this equation is independent from the gate and calculated by
\begin{equation}
    \frac{\partial z_k^t}{\partial c_k^t} = \frac{\partial}{\partial c_k^t} \big(o_k^t 
    \textrm{tanh}\big(c_k^t\big)\big) = o_k^t \textrm{tanh}'\big(c_k^t\big) .
\end{equation}
The computation of an eligibility value $\epsilon_{jk}^{(A,B),t}$ is described by
\begin{equation}
\label{eq:evaluelstm}
    \epsilon_{jk}^{(A,B),t} = \epsilon_{jk}^{(A,B),t-1} \frac{\partial c^t_k}{\partial c^{t-1}_k} + \frac{\partial c^t_k}{\partial w_{jk}^{(A,B)}},
\end{equation}
where $\frac{\partial c^t_k}{\partial c^{t-1}_k}$ resembles the internal dynamics of the network, which can be modeled by the forget gate $f^{t}_k$ for LSTMs \cite{bellec2019biologically,bellec2020solution}. The crucial part for the computation of an eligibility value is $\frac{\partial c^t_k}{\partial w_{jk}^{(A,B)}}$. This derivative depends on the specific weights of a gate and therefore has to be calculated individually for input and recurrent weights as well as the bias of the gates. 

As an example, we derive the eligibility trace for the input weights of the input gate. The crucial derivative is described by
\begin{equation}
        \frac{\partial c^t_k}{\partial w_{jk}^{(i,\textrm{in})}} = \tilde{c}_k^t i_k^t(1-i_k^t) x_j^{t} ,
    \end{equation}
which leads to the complete expression of the eligibility trace $e^{(i,\textit{in}),t}_{jk}$ being
\begin{equation}
    e^{(i,\textrm{in}),t}_{jk} = \underbrace{\big( \epsilon_{jk}^{(i,\textrm{in}),t-1} f_k^t + \underbrace{\tilde{c}^t_k \sigma'\big(net_k^{(i),t}\big) x_j^t\big)}_{\frac{\partial c^t_k}{\partial w_{jk}^{(i,\textrm{in})}}}}_{\epsilon_{jk}^{(i,\textrm{in}),t}} \underbrace{o_t^j \textrm{tanh}'\big(c_k^t\big)}_{\frac{\partial z_k^t}{\partial c_k^t}} .
\end{equation}

The output gate contributes only to the hidden state $h_k^t$ of an LSTM. Since we defined $c_k^t$ as the hidden variable for e-prop, we do not need to calculate eligibility traces to compute gradients for the output gates. Instead, we can use the same factorization of the error gradient as in BPTT. The gradient of the error with respect to $w_{jk}^{(o,\textrm{in})}$ is therefore expressed as
\begin{equation}
    \frac{d E}{dw_{jk}^{(o, \textrm{in})}} = \sum_t \frac{dE}{dz_k^t} \frac{\partial dz_k^t}{\partial w_{jk}^{(o, \textrm{in})}} = \sum_t \frac{dE}{dz_k^t}\textrm{tanh}'(c_k^t)o^t_k(1-o_k^t)x_j^t .
\end{equation}

\subsubsection{Learning Signals}
\label{subsection:learningsignalslstm}

According to \cite{bellec2019biologically,bellec2020solution} a learning signal at timestep $t$ is given by 
\begin{equation}
    l_k^t = \sum_n m_{kn}(y_n^t - y_n^{*,t}) ,
\end{equation}
where $y_n^t$ is the current output of the network for the neuron $n$ and $y_n^{*,t}$ is the target. The value $m_{kn}$ is determined by a feedback matrix $M$. We consider two versions of e-prop to define $M$. For random e-prop, a random matrix, initialized in the same way as the network's weights, is used. For symmetric e-prop, the transpose of the weights to the output layer is chosen $M = W^{\top}$. This can be straightforwardly applied to LSTMs. Note that the output layer is updated via plain backpropagation since it is not involved in any recurrent computations.

Depending on the problem, there are two options how learning signals can be used -- either at all timesteps or only at the final timestep. Therefore, gradients can be computed either by accumulating the combination of eligibility traces and learning signals over time, as described in equation \ref{equation:gradient_all_timesteps},
or by using only the combination of the final eligibility trace and the final learning signal
\begin{equation}
    \frac{dE}{dw_{jk}^{(A,B)}} = e_{jk}^{(A,B),T} l_k^T.
\end{equation}

\subsection{Extensions to LSTM based E-Prop}

\subsubsection{Initialization of the Forget Gate Bias}

The idea of initializing the forget gate bias with higher values was already discussed at the introduction of forget gates \cite{lstm_forget} and later work proved this can boost training significantly \cite{jozefowicz2015empirical}. Doing so is a change to the architecture and can also be applied when training LSTMs with BPTT. However, e-prop will profit differently from this due to the special role of the forget gate on which e-prop highly depends. As seen in equation \ref{eq:evaluelstm}, the influence of previous eligibility values on current ones depends exclusively on the forget gate. We can make use of this by initializing the bias of this gate with a higher value, which increases the influence of the previous eligibility value on the current one in e-prop during the initial training phase.

\subsubsection{Trace Echo}

Using only the final eligibility trace and learning signal can be problematic for longer sequences or sequences where most of the information lies in the first timesteps since the eligibility trace could fade over time. To tackle this specific problem, we introduce the trace echo. 

The trace echo is a modulated signal of all eligibility traces during a sequence accumulated over time decoupled from the (supervised) learning signal. It resembles the overall influence of the networks weights on the output of the postsynaptic neuron given presynaptic activity over the whole sequence. The trace echo term can thus be seen as an additional unsupervised learning component, loose implementing a form of Hebbian-learning \cite{hebb1949organization}.

In order to apply the trace echo, we add a weighted sum of all eligibility traces over time to the gradient computation:
\begin{equation}
    \frac{dE}{dw_{jk}} = e_{jk}^T l_k^T + \underbrace{\lambda \sum_{t=0}^T e_{jk}^t}_{\textrm{trace echo}} .
\end{equation}
Of course, the trace echo is neither limited to learning with only the final timestep nor to LSTM based e-prop.

\subsubsection{Trace Scaling}

During experiments with DRQN, we discovered that in the beginning of training, the computation of eligibility values can be highly unstable. This can lead to suboptimal weight changes which make further learning impossible. To address this problem, we introduce trace scaling.

Trace scaling heavily scales down the eligibility traces, and therefore their influence on weight updates, by introducing a factor $\mu$ to their computation,
\begin{equation}
    e^{(A,B),t}_{jk} = \mu_{B} \epsilon_{jk}^{(A,B),t} \frac{\partial z_k^t}{\partial c_k^t} .
\end{equation}

Trace scaling also shifts the focus for weight updates more towards the output layer since the updates for this layer are not computed via e-prop. This can be advantageous when there is an imbalance in the updates of the two layers. Learning with trace scaling will be slower for the recurrent layer and the output layer can learn to adapt to the recurrent layer, rather than simply counterbalancing the output of the recurrent layer. It should be noted that optimizers with adaptive learning rates can prevent this. Trace scaling is neither limited to LSTM based e-prop nor to DRQN.

\subsection{E-Prop in the Context of Reinforcement Learning}

DRQN \cite{hausknecht2015deep} is an approach, able to deal with POMDP, that uses LSTMs to estimate Q-values. LSTMs are able to learn to internally store relevant information about previous timesteps, which can then be used in addition to current observations to choose the best possible action. Training is done with BPTT on sequences of experiences. These sequences are sampled at a fixed and rather short length from the episodes an agent performs on an environment. This is done for two reasons. Firstly, we need to store the experience-sequences for later replay and shorter sequences take up less space. Secondly, and even more importantly, it is impractical to use long sequences when training with BPTT.
Because e-prop can work forward in time, the network no longer has to be unrolled, putting e-prop into an advantageous position for RL. For this work we use DRQN with the inclusion of actions to the network input \cite{zhu2017improving}. To our knowledge, the integration of e-prop to recurrent Q-learning is novel. Integrating e-prop to DRQN is therefore only supposed to be a proof of concept for this approach.

\section{Experiments and Results}

\subsection{Classification of Handwritten Digits}
Image classification tasks like MNIST are typically learned feedforward with convolutional neural networks, but can also be learned sequentially. For sequential MNIST (sMNIST) the images are transformed into a sequence of single pixels going from the upper left corner to the bottom right one. This results in a sequence with a length of 784 pixels. Learning MNIST in sequential manner provides one with a quite difficult temporal credit assignment problem on long sequences, where information is sparse. Permuting the order of the pixels by a fixed permutation \cite{psMNIST} yields a harder version of sMNIST, where the information is still sparse but not clustered like before.

We established a BPTT baseline with an architecture of 128 LSTM units and one output layer with a softmax activation function based on \cite{arjovsky2016unitary} for both tasks. In contrast to \cite{arjovsky2016unitary}, we found that a lower learning rate of 0.0005 for RMSProp \cite{rmsporp} and no gradient clipping yields better results for BPTT.

\subsubsection{sMNIST Results}
E-prop worked best with a learning rate of 0.001. We performed five training runs with both symmetric and random e-prop, where a label was presented at the final timestep only. Furthermore we performed runs with a higher initialized forget gate bias, the trace echo extension and a combination of both. The forget bias was initialized as the other weights, before increasing it by 1.8. For the trace echo $\lambda = \frac{1}{\textrm{sequence length}} 10^{-4}$ was used. The final results are given in table \ref{table:sMNIST_results}.

\begin{table}[b!]
\centering
\caption{Results for sMNIST. Given is the mean accuracy of five runs on the test set after 350 epochs.}
\begin{tabularx}{\linewidth}{ccCCC} 
\toprule
algorithm & ~~~~~~~~plain~~~~~~~~ & forget gate bias & trace echo & combined\\
\midrule 
random e-prop & 88.77\% & 58.30\% & 65.99\% & 70.69\% \\
symmetric e-prop & 92.76\% & 93.49\% & 91.37\% & 92.49\% \\
BPTT & 98.92\% & -- & -- & --\\
\bottomrule
\end{tabularx}
\label{table:sMNIST_results}
\end{table}

\begin{figure}[t!]
\centering
\includegraphics[width=\linewidth]{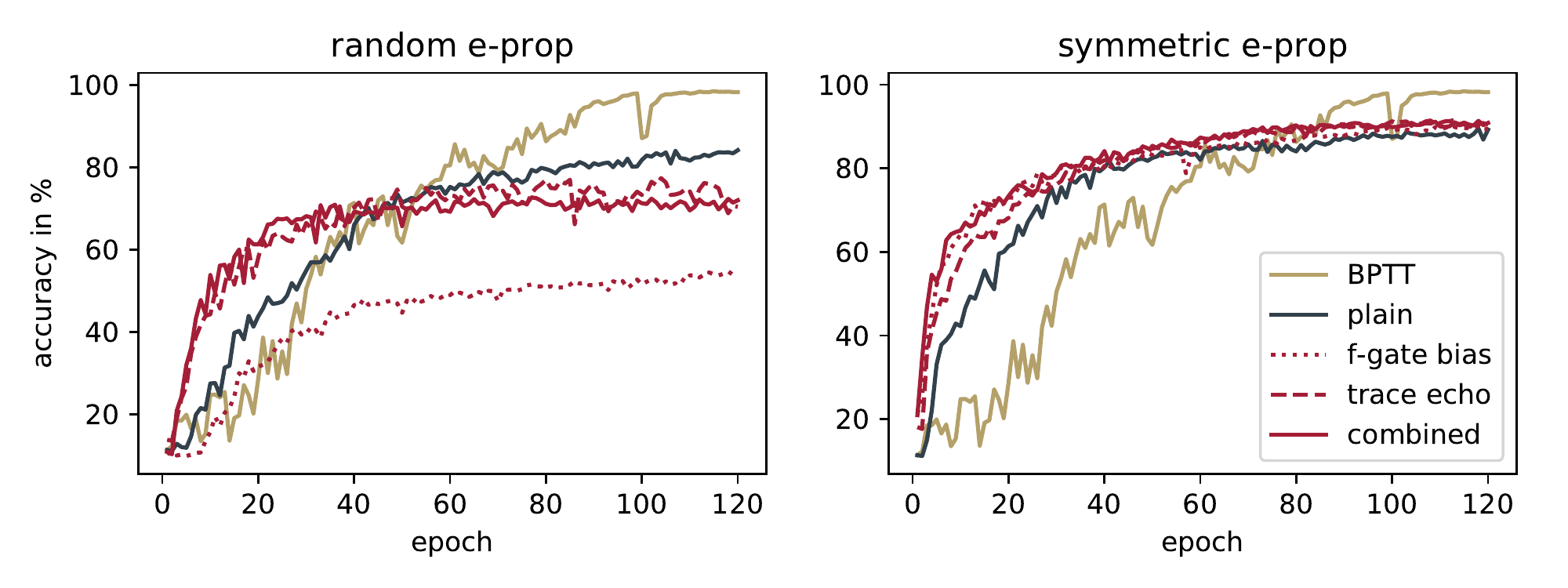}
\caption{Comparison of e-prop extensions and BPTT on sMNIST. Depicted is the mean accuracy of five runs on the test set for both random e-prop (left) and symmetric e-prop (right) for the first 120 epochs of training. A BPTT baseline is given for both.}
\label{fig:sMNIST_ev_extensions}
\end{figure}

During the first 100 epochs, the extensions had a negative impact on random e-prop, training was unstable and multiple runs broke during training. Symmetric e-prop was improved considerably during the first epochs. However, this improvement is not persistent and the final results after 350 epochs are comparable to the plain version. The influence of the extensions during the first 120 epochs of training is visualized in figure \ref{fig:sMNIST_ev_extensions}.

\subsubsection{Permuted sMNIST Results}
\begin{figure}[t!]
\centering
\includegraphics[width=\linewidth]{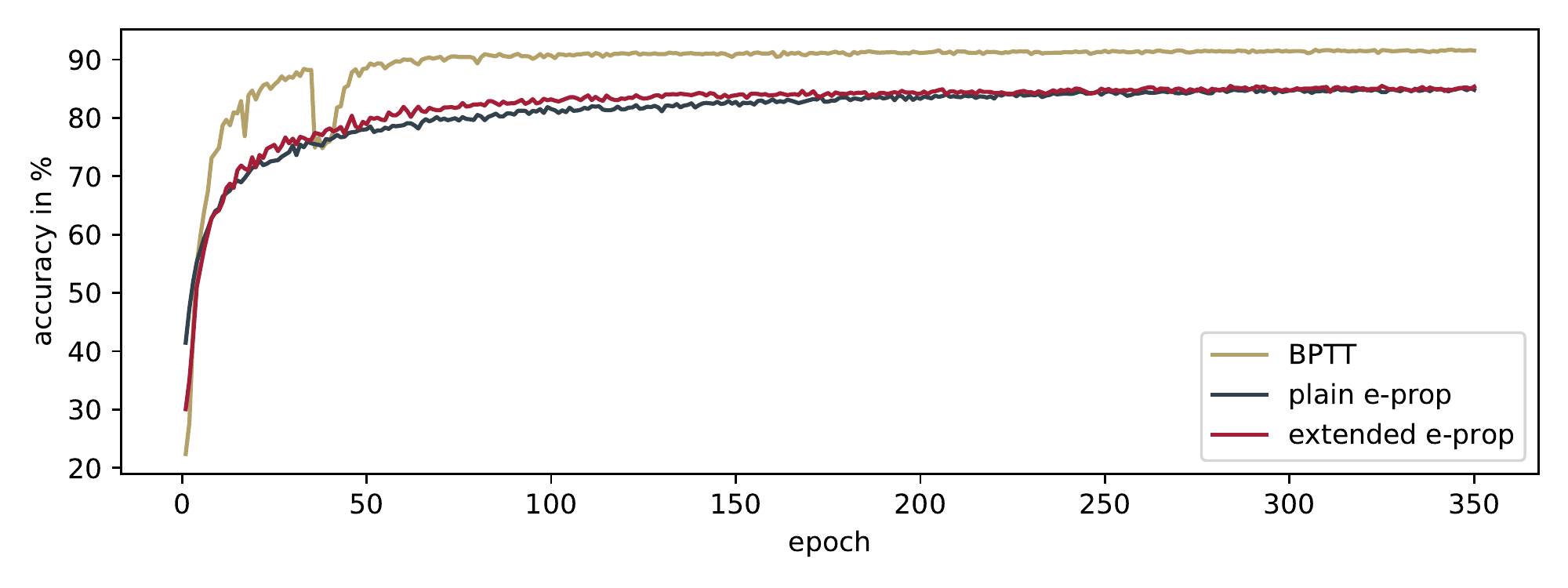}
\caption{BPTT vs. symmetric e-prop on permuted sMNIST. Given is the mean accuracy on the test set during training. BPTT performs better than symmetric e-prop. Performance of e-prop can be improved by trace echo and higher forget gate bias for the beginning of training.}
\label{fig:psMNIST_all}
\end{figure}

BPTT reached a mean accuracy of 91.55\% on the test set after 350 epochs over five runs. The saturation of training was reached earlier then previously. Plain symmetric e-prop reached a mean accuracy of 84.83\%. An effect of combining trace echo and higher forget gate bias initialization is less pronounced than previously. Extended e-prop archives a final mean accuracy of 85.36 \%. The results are depicted in figure \ref{fig:psMNIST_all}.

\subsection{Solving another Temporal Credit Assignment Problem}

\subsubsection{TCA Task}

To further investigate temporal credit assignment (TCA) we introduce another task, where sequences consists of four different channels. The first two channels produce random values between zero and one during the whole sequence. The third one indicates, with a probability of 0.5, whether the information of the first two channels is valuable (one) or not (zero). After 15 timesteps, the channel remains zero for a certain delay. The last channel produces zeros until the last timestep, where it switches to one. A possible sequence could be:
\setcounter{MaxMatrixCols}{20}
\begin{equation*}
\begin{matrix}
\text{\textbf{left}:}\\
\text{\textbf{right}:}\\
\text{\textbf{cue}:}\\
\text{\textbf{done}:}
\end{matrix}
\begin{bmatrix}
0.42 & 0.49 & 0.35 & 0.76 & 0.69 & 0.19 & 0.99 & \ldots & 0.85 & 0.23 & 0.42\\
0.75 & 0.67 & 0.86 & 0.57 & 0.37 & 0.92 & 0.55 & \ldots & 0.47 & 0.28 & 0.40\\
0 & 1 & 0 & 1 & 1 & 0 & 0 & ... & 0 & 0 & 0\\
0 & 0 & 0 & 0 & 0 & 0 & 0 & ... & 0 & 0 & 1
\end{bmatrix}
\end{equation*}

A network has to compare the value of the first two channels whenever the third is active. The side with the higher value wins. In the end, a decision, which side has won more often, has to be made. The network, therefore, has to learn when it should listen, to compare two numbers, to count the number of wins for each side, to determine a winning side, and to store this information over time.

New batches of data were generated for all update steps. This ensures that networks learn to generalize. A test or validation set is, therefore, not needed, the running mean of the training error is sufficient for an analysis.

\subsubsection{Results}

A network with 32 LSTM cells, followed by an output layer, was used. The hyper-parameters of the e-prop extensions were chosen as previously. 90,000 updates on batches with a size of 64 were performed with a delay of 10, 15, 20 and a random value between 10 and 20 timesteps. We performed 10 different runs for a extended version of symmetric e-prop with higher initialized forget gate bias and trace echo, and for BPTT. 

\begin{figure}[t!]
\centering
\includegraphics[width=\linewidth]{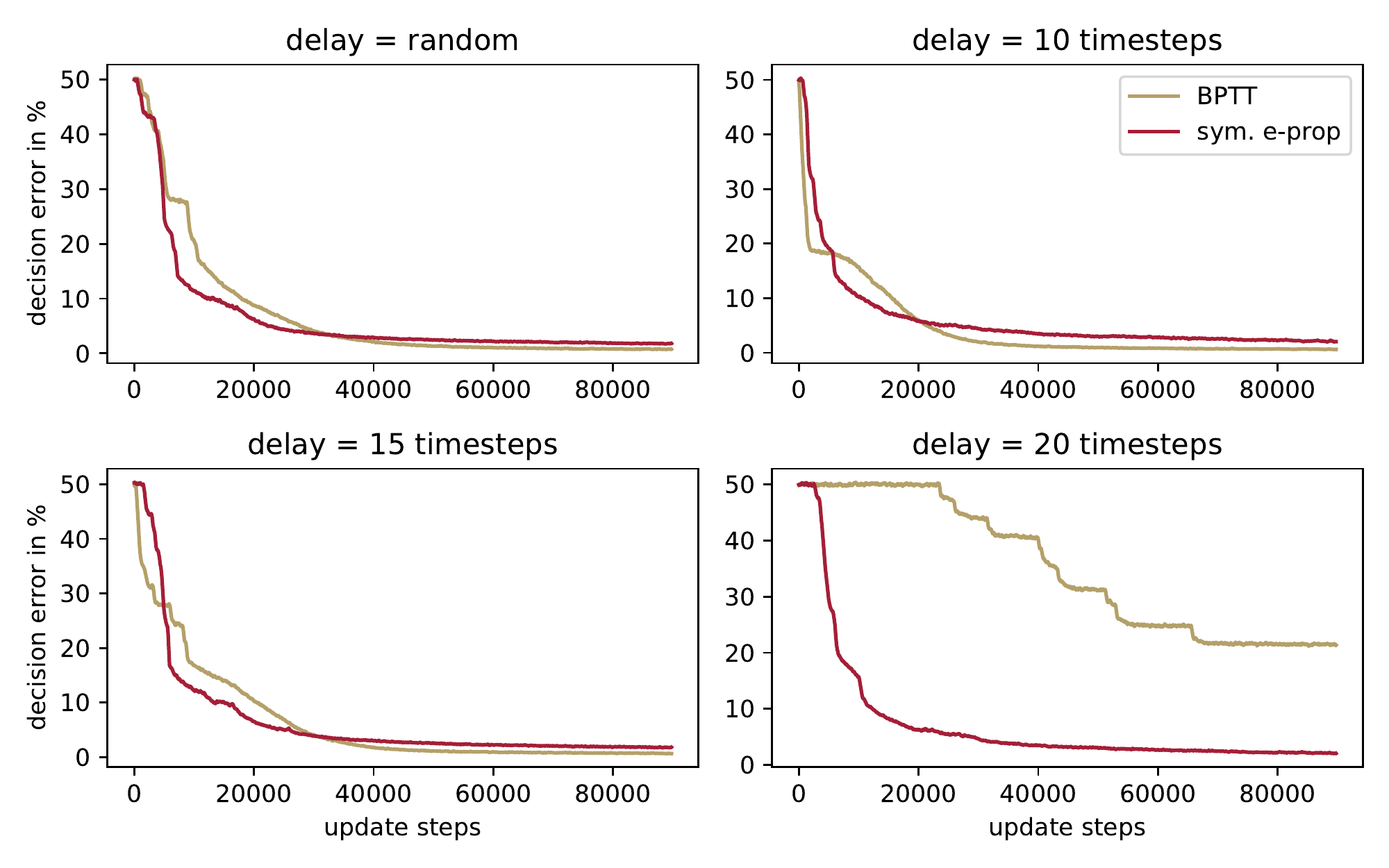}
\caption{Training of BPTT and extended symmetric e-prop on the TCA task, mean of 10 runs. Shown is the running mean over 250 update steps of the decision error of a batch during training. E-prop was able to learn the task independently from the delay condition. BPTT failed to master the task in the condition with the largest delay and is outperformed by e-prop.}
\label{fig:tca_results}
\end{figure}

RMSProp worked best for e-prop and Adam \cite{kingma2014adam} for BPTT except for the 20 timestep delay condition, where RMSProp worked better. Extended e-prop was trained with a learning rate of 0.01. Training with an increased learning rate failed for both plain e-prop and BPTT. Figure \ref{fig:tca_results} shows the training of BPTT and extended e-prop for all conditions, the final results are given in table \ref{table:TCA_results_90k}.

\begin{table}[b!]
\centering
\caption{Results for the TCA task after 90,000 update steps. Given is the decision error for BPTT and e-prop with trace echo and a higher initialized forget gate bias.}
\begin{tabularx}{\linewidth}{CCC}
\toprule
delay & extended e-prop & BPTT\\
\midrule 
random & 1.79\% & 0.72\%\\
10 & 2.03\% & 0.62\%\\
15 & 1.79\% & 0.64\%\\
20 & 2.02\% & 21.37\%\\
\bottomrule
\end{tabularx}
\label{table:TCA_results_90k}
\end{table}

\begin{figure}[t!]
\centering
\includegraphics[width=\linewidth]{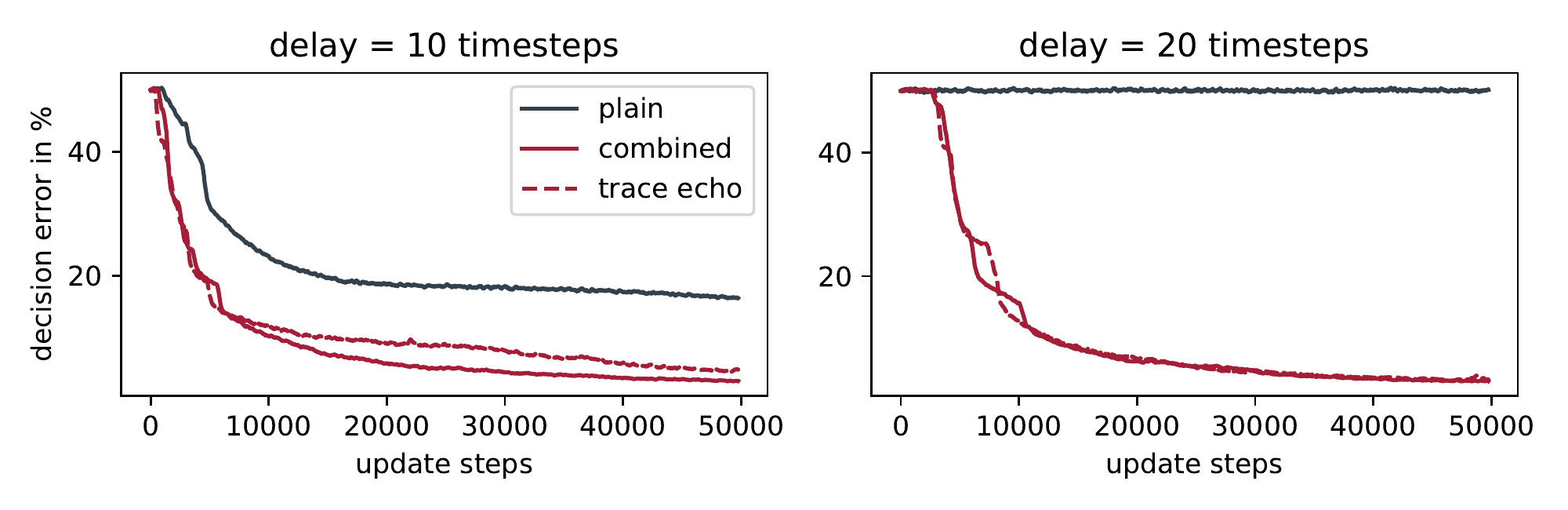}
\caption{Training of e-prop versions on the TCA task, mean of 10 runs. Shown is the running mean over 250 update steps of the decision error of a batch during training. The trace echo boosts training considerably}
\label{fig:tca_1_20_comp}
\end{figure}

To ensure that the extended e-prop version did not perform better due to the initialization of the forget gate bias alone, we performed 10 additional runs with 50,000 updates on the conditions with a delay of 10 and 20 timesteps. The results are depicted in figure \ref{fig:tca_1_20_comp} as well as the results of plain e-prop for these conditions. It should be noted that reliable training with plain e-prop was only possible in the condition with a delay of 10 timesteps. The final results are given in table \ref{table:TCA_results_50k}.  

\begin{table}[t!]
\centering
\caption{Results for the TCA task after 50,000 updates. Given is the decision error for BPTT and the different e-prop variations used for the easiest and hardest delay condition.}
\begin{tabularx}{\linewidth}{XXXXX}
\toprule
~ & ~ & \multicolumn{3}{c}{e-prop}\\ \cmidrule{3-5}
delay & BPTT & plain & trace echo & combined\\
\midrule
10 & 0.94\% & 16.35\% & 4.82\% & 3.0\%\\
20 & 31.19\% & 50.10\% & 3.32\% & 3.07\%\\
\bottomrule
\end{tabularx}
\label{table:TCA_results_50k}
\end{table}

\section{Solving Easy RL Environments with DRQN}

\subsection{Environment}

CartPole-v0 was chosen as an environment because it is fairly easy to learn and, therefore, suited for a proof of concept. A network with 32 LSTM units and an output layer was used with Adam (lr=0.0003). For the e-prop based version, a high value for $\epsilon_{\textrm{Adam}} = 0.1$ proved to be essential to solve the environment. For trace scaling we used $\mu_{\textrm{in}}=10^{-4}$ and $\mu_{\textrm{rec,bias}}=10^{-8}$ . To fill the replay buffer, 200 episodes of random actions were performed at the beginning of training.

Inspired by flickering Atari games \cite{hausknecht2015deep}, we present a partially observable version of CartPole-v0, by zeroing out the observations with a probability of $0.2$.

\subsubsection{CartPole Results}

\begin{figure}[t!]
\centering
\includegraphics[width=\linewidth]{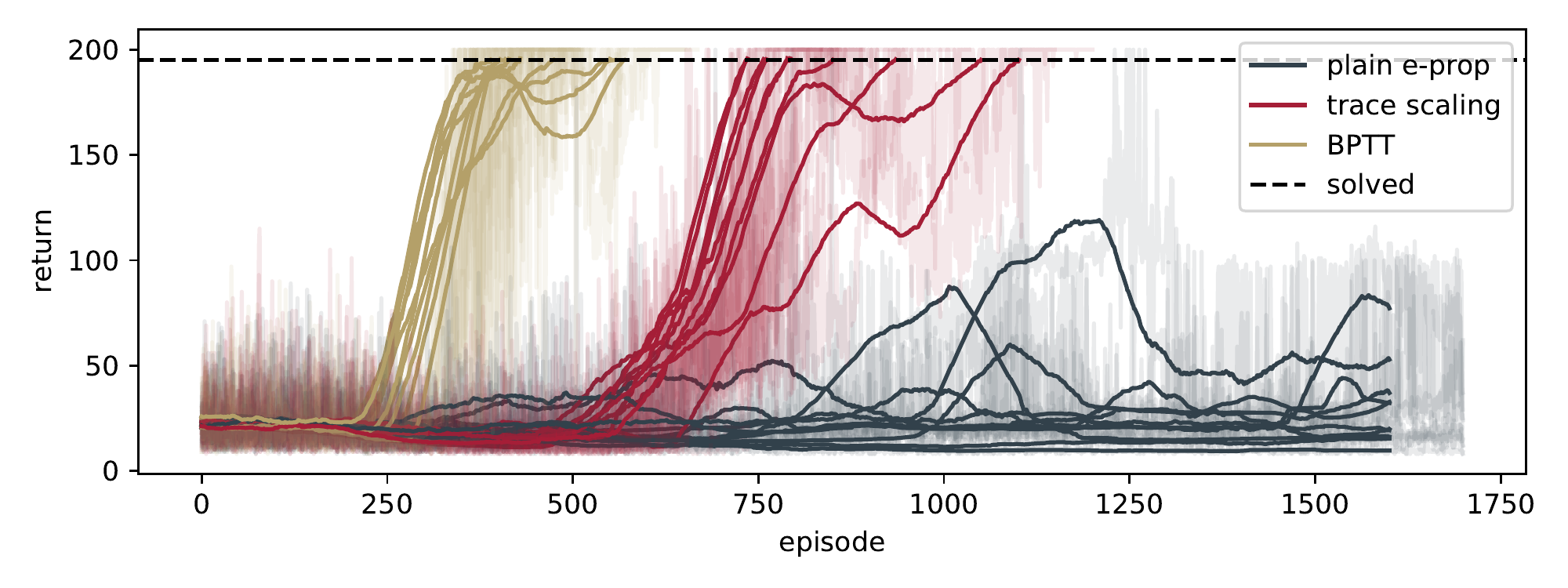}
\caption{Training of CartPole-v0 with BPTT and e-prop. Shown are 10 training runs for BPTT, e-prop with trace scaling and plain e-prop. The running mean of the returns over the last 100 episodes is highlighted. BPTT learns to solve the environment significantly faster than e-prop. Training with plain e-prop was not achieved.}
\label{fig:CP_BPTTvsEPROP}
\end{figure}

Solving the environment with e-prop based DRQN was made possible by trace scaling. However in a direct comparison with BPTT, e-prop proved to be inferior. BPTT based e-prop was able to solve the environment significantly faster for all runs we performed. Training is depicted in figure \ref{fig:CP_BPTTvsEPROP}.

\subsubsection{Partially Observable CartPole Results}

\begin{figure}[t!]
\centering
\includegraphics[width=\linewidth]{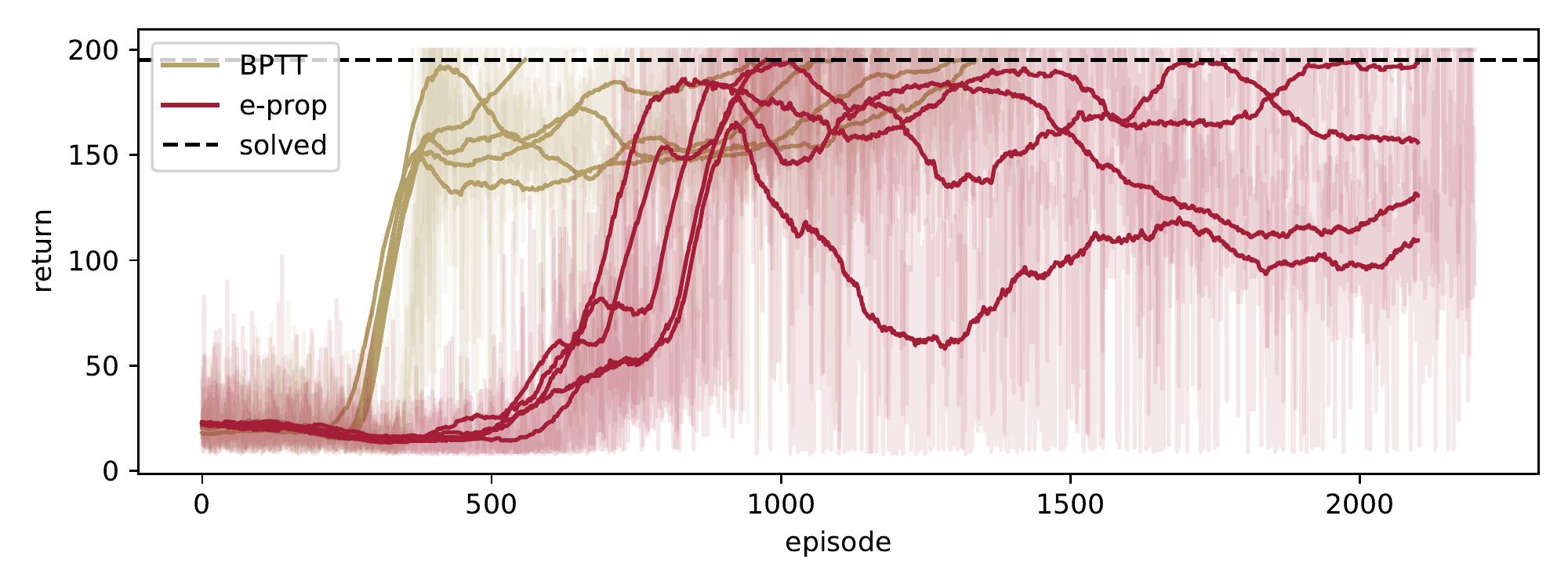}
\caption{Training of a partially observable CartPole-v0 with BPTT and e-prop. Shown are the returns of five training runs smoothed over the last 100 episodes for BPTT and e-prop with trace scaling.}
\label{fig:pocp}
\end{figure}

Training takes considerably longer, showing the increased difficulty of the environment. Training with BPTT starts comparable to standard CartPole until a mean return of around 150 is reached. All five runs with BPTT solved the environment. Only one out of five e-prop runs was successful, the others show learning, and achieved mean returns of over 150 but failed to solve the environment. The results can be observed in figure \ref{fig:pocp}. Runs without trace scaling failed again.

\section{Discussion}

\subsubsection{sMNIST}
Our results prove that the forward computation of eligibility traces is stable for sequences of several hundred timesteps, and that the final eligibility trace is sufficient for the computation of weight updates. Both random and symmetric e-prop were able to learn the sequential MNIST task but ultimately achieved lower accuracies than BPTT. Random e-prop turned out to be inferior to symmetric e-prop, which is no surprise since symmetric e-prop works with better learning signals, computed with the true backward connections.

We showed that our extensions can improve training with symmetric e-prop. A combination of trace echo and a higher initialized forget gate bias stabilized training during the initial training phase considerably. Both of these extensions on their own boosted training during the initial phase as well. Significantly improved results were not achieved with our extensions after full training.

Compared to BPTT, the results for e-prop are not satisfactory. The initial training phases of both random and symmetric e-prop look considerably smoother than the training of BPTT. However, BPTT reaches a significantly higher accuracy than e-prop.

\subsubsection{TCA}

The trace echo in combination with a biased forget gate allowed symmetric e-prop to learn the TCA task. The performance of e-prop was drastically increased by the extensions. Learning TCA with plain e-prop was not possible for most conditions.

Extended e-prop outperforms BPTT by far in the hardest condition and performs comparable in the others. By replicating these results with only the trace echo, we show that this success is due to the trace echo and not due to the changed initialization of the forget gate bias alone, which could boost BPTT in theory as well.

The discovery that the trace echo can improve e-prop in such a drastic manner is a major contribution. With the trace echo extension, we introduce an unsupervised learning signal that resembles a sort of Hebbian learning to e-prop, which differs from the type of learning performed by plain e-prop and BPTT.

\subsubsection{DRQN}

Trace scaling proved to be indispensable for achieving any learning with e-prop based DRQN. Even though it allowed reliable training with e-prop based DRQN, the performance is not comparable to BPTT based DRQN, which solved the environment considerably faster. This is partially due to higher values of $\epsilon_{\textrm{Adam}}$ for e-prop runs, however, without a smaller $\epsilon_{\textrm{Adam}}$, the environment was not solved. Learning with e-prop based DRQN was also achieved for our partially observable version of CartPole-v0. But, unlike BPTT, e-prop based DRQN was not able to solve this environment reliably.

The main problem that e-prop has to face in DRQN is that the targets for the LSTM are only estimates based on the ever-increasing experience of an agent. The better an agent gets, the higher the accumulated returns and targets become. The result of these ever-increasing targets is, in contrast to supervised learning, that the loss can indeed rise during training when the agent is performing better. The fact that this estimation of Q-values diverges and is unstable is one of the main problems of DQN approaches. Using a less stable optimization algorithm like e-prop increases this problem and is most likely the cause for our problems with the CartPole environments. Future works with e-prop based DRQN should include already known DQN extensions.

\section{Conclusion}
In this work, we covered the application of e-prop to LSTMs. Our results show that symmetric e-prop is a highly capable algorithm for the optimization of LSTMs. The introduced extensions were able to further improve symmetric e-prop. The trace echo extension proved to be particularly valuable. It includes an unsupervised learning signal based on past eligibility traces and clearly alleviates some of the limitations of e-prop. We showed that under certain conditions, symmetric e-prop with the trace echo can outperform BPTT.
Furthermore, we delivered a proof of concept for the application of e-prop to recurrent Q-Learning.

\bibliographystyle{splncs04}
\bibliography{2022-ExtendedEProp}
\end{document}